\documentclass{opt2024} 

\usepackage{csquotes}
\usepackage{ifthen}
\usepackage{enumitem}
\usepackage{amsmath,amssymb,amsfonts}
\usepackage{bm}
\usepackage{graphicx}
\usepackage{xspace}
\usepackage{verbatim}
\usepackage{algorithm}
\usepackage{algpseudocode}

\usepackage{color}
\usepackage{thm-restate}
\usepackage{latexsym}

\usepackage[colorinlistoftodos,textsize=scriptsize]{todonotes}

\usepackage{fancybox} 
\usepackage{tcolorbox}

\title[Neural Networks with Complex-Valued Weights Have No Spurious Local Minima]{Neural Networks with Complex-Valued Weights \\ Have No Spurious Local Minima}

\optauthor{%
\Name{Xingtu Liu} \Email{rltheory@outlook.com}\\
\addr Simon Fraser University}

\begin{document}

\maketitle

\begin{abstract}%
We study the benefits of complex-valued weights for neural networks. We prove that shallow complex neural networks with quadratic activations have no spurious local minima. In contrast, shallow real neural networks with quadratic activations have infinitely many spurious local minima under the same conditions. In addition, we provide specific examples to demonstrate that complex-valued weights turn poor local minima into saddle points.
\end{abstract}


\section{Introduction}
A major challenge in deep learning is to avoid gradient descent from getting trapped in suboptimal local minima. Thus, understanding the optimization landscape of neural networks has become a significant area of research. It has been shown that almost all non-linear real-valued neural networks have poor local minima, which includes neural networks with {\it quadratic}, {\it tanh}, {\it sigmoid}, {\it arctan}, ReLU, ELU, and SELU activations \cite{yun2018small}. Even under the over-parametrization condition, real-valued neural networks have strict poor local minima \cite{9194023}. Only linear neural networks have been proven to have no spurious local minima \cite{NIPS2016_6112}, however, they are rarely used in practice because of their limited expressive power. Universal approximation theorem only holds for \textit{non-linear} neural networks.

In this paper, we prove a simple but surprising result:
\begin{center}
    \begin{tcolorbox}[colback=white,colframe=black,boxrule=0.5mm]
    \centering
        \textit{All local minima in shallow complex neural networks with quadratic activations \\ are global minima.}
    \end{tcolorbox}
\end{center}
This result is unanticipated for the following reasons.
\begin{itemize}
    \item Under the exact same conditions (i.e., shallow neural networks with quadratic activations), real-valued networks have infinitely many spurious local minima.
    \item No unrealistic assumptions are made, and not even mild over-parameterization (i.e., $k > n$, where $k$ is the number of hidden nodes and $n$ is the number of samples) is required .
\end{itemize}
We only require that $k \geq d$, where $d$ is the dimension of each sample. This implies that the number of hidden nodes is at least equal to the \textit{dimension of each sample}, which is a common setting. In contrast, the assumption of mild over-parametrization requires the number of hidden nodes to be larger than the \textit{number of input samples}, i.e. $k \geq n$. Moreover, the standard over-parametrization setting, for example in neural tangent kernel (NTK), the assumption of $k \geq O(n^4)$ was made.

In addition, we provide some specific examples of poor local minima in the real-valued case and demonstrate how they become saddle points in the complex-valued neural networks (CVNNs).

Experimental results have shown that complex-valued weights are more likely to escape poor local minima in neural networks with polynomial activations \cite{bubeck2020network, pmlr-v32-andoni14}. However, this work is the first to theoretically demonstrate that the loss surface of nonlinear complex-valued neural networks is indeed superior to that of their real-valued counterparts. The minimum modulus principle in complex analysis tells us complex analytic functions have superior landscapes. Therefore, we conjecture that the \enquote{no spurious local minima} result also holds true for deeper complex networks and complex networks with other analytic activations such as {\it tanh} and {\it sigmoid}. Along the way, we develop a novel set of tools and techniques for analyzing the optimization landscape of CVNNs, which may be useful in other contexts. 

\subsection{Related Work}

The analysis of optimization landscape of neural networks began with linear neural networks. Baldi et al. \cite{BPK89, BL12} proved the \enquote{no spurious local minima} result for both real-valued and complex-valued shallow linear neural networks. This was later translated to deep linear neural networks by Kawaguchi \cite{NIPS2016_6112}. Work on the optimization of linear networks from other perspectives can be found elsewhere \cite{hardt2018identity,lu2017depth,pmlr-v80-laurent18a,zhang2020depth}. Although optimizing linear networks being a non-convex problem have a nice landscape, the representation power of linear network is limited as they can only fit linearly separable data. 

For non-linear networks, all demonstrations of \enquote{no spurious local minima} required unrealistic assumptions or an extremely restricted parameter space. For instance, independence between weights was assumed for deep ReLU networks \cite{NIPS2016_6112}. For shallow ReLU networks, the result in \cite{wu2018no} only holds for two hidden units networks and weight vectors must to be unit-normed and orthogonal. For shallow quadratic networks, \cite{NIPS2019_9111} assumed Gaussian feature vectors; while \cite{SJL19} assumed the weight vector connecting the hidden layer and the output node must contain at least d positive entries and d negative entries. Without these unrealistic assumptions, all common non-linear real-valued networks were shown to have poor local minima \cite{yun2018small}. 

The optimization landscape of neural networks under over-parametrization has also been studied \cite{pmlr-v70-nguyen17a,nguyen2018on, li2021benefit}. Unfortunately, it has been shown that neural networks with over-parametrization can still have strict poor local minima \cite{Sharifnassab2020Bounds, 9194023}, and utilizing the set of measure zero only eliminated non-strict poor local minima, i.e. sub-optimal basins. 

Some attempts from the algorithm dynamics have been made, showing that gradient descent can converge to global minima under certain conditions \cite{du2019gradient, pmlr-v97-du19c}. However, the networks in the analysis are ultra-wide.

\section{Preliminaries}
\label{sec_pre}

This section provides an introduction to complex analysis and Wirtinger calculus. More definitions and lemmas are provided in the appendix.

\subsection{Notations}

Let $\mathbb{R}$ and $\mathbb{C}$ denote the real and complex fields respectively. They share many common properties as number fields. Note that $\mathbb{R} \subseteq \mathbb{C}$ and $\mathbb{R}^{m \times n} \subseteq \mathbb{C}^{m \times n}$.  Let $z = z_1 + iz_2 \in \mathbb{C}$, we use $||z || = \sqrt{z_1^2 + z_2^2} \in \mathbb{R}$ to denote its modulus and $z^* = z_1 - iz_2$ to denote its conjugate.  $\mathcal{R}(\textbf{z})$ and  $\mathcal{I}(\textbf{z})$ are used to denote the real and imaginary part of a complex vector $\textbf{z} \in \mathbb{C}^n$. For $\textbf{M} \in \mathbb{C}^{m \times n}$, $\textbf{M}^T$ and $\textbf{M}^*$ are the transpose and conjugate transpose of $\textbf{M}$, $\textbf{M}^C$ denotes the matrix whose entries are conjugates of entries in $\textbf{M}$, Null$(\textbf{M}) := \{\textbf{v} \ |\  \textbf{M}\textbf{v}=0 \}$ denotes the null space, and vec$(\textbf{M})$ denotes the vectorization of $\textbf{M}$. For $z \in \mathbb{C}$ and $\textbf{M} \in \mathbb{C}^{m \times n}$, we have $z^* = z^C$ and $\textbf{M}^* = (\textbf{M}^C)^T$.

\subsection{Complex functions}
A complex function $f: \mathbb{C} \mapsto \mathbb{C}$ is given by $f(z) = u(z) + iv(z)$. We can also think of $f$ as $f: \mathbb{R}^2 \mapsto \mathbb{R}^2$ where $f(x,y) = (u(x,y),v(x,y))$. A complex-valued multivariate function $f : \mathbb{C}^n \mapsto \mathbb{C}$ is given by $f(\textbf{z}) = u(\textbf{z}) + iv(\textbf{z})$ where $u(\textbf{z}), v(\textbf{z}) \in \mathbb{R}$. The most prominent function used in this work is a real-valued function with complex-valued matrix input $f : \mathbb{C}^{m\times n} \mapsto \mathbb{R}$.

A complex function is analytic if it is differentiable at every point and in it's neighborhood in the domain. An analytic function must satisfy the Cauchy Riemann equations (CRE). See the appendix for the definitions of differentiable, analytic, CRE, and more. Note that a non-constant real-valued complex function does not satisfy the CRE and is thus not analytic.

Recall that the loss function
$$
\mathcal{L}(\textbf{W}) = \frac{1}{2n}\sum^n_{i=1}\parallel y_i - \textbf{v}^T\psi(\textbf{W}\textbf{x}_i)\parallel^2
$$
is not analytic. However, $y_i - \textbf{v}^T\psi(\textbf{W}\textbf{x}_i)$ is analytic for each $i$, which indicates that their derivatives are well defined.

\subsection{Wirtinger calculus}

Since $\mathcal{L}(\textbf{W})$ is not differentiable in the traditional sense, we require a new way of calculating the complex gradient. Many non-differentiable complex functions are differentiable in the real sense if we treat $ \mathbb{C}^n $ as $\mathbb{R}^{2n}$. $\mathcal{L}(\textbf{W})$ is one of them. Wirtinger calculus, also known as the $\mathbb{C}\mathbb{R}$-Calculus, provides a neat way of deriving the derivatives. For a differentiable complex function, Wirtinger derivatives are the same as the traditional derivatives. Using Wirtinger calculus provides not only defined derivatives that reflect a function's gradient, but also have meaningful results for the first and second derivatives. Note that critical points, positive semi-definite Hessian, and Taylor expansions all have their counterparts in Wirtinger calculus. Due to space constraints, an extended introduction of Wirtinger calculus can be found in Appendix. A comprehensive introduction can be found in \cite{KreutzDelgado2005TheCG} and \cite{BP10}.

\section{Main Results}

We analyze shallow neural networks with {quadratic} activation functions over the fields $\mathbb{R}$ and $\mathbb{C}$, with the network structure and loss function defined as below.

\begin{definition}
\label{def_main}
Define the dataset $\{(\textbf{x}_i, y_i)\}$ for $i=1,2,\dots, n$ with $\textbf{x}_i \in \mathbb{F}^d$ and $y_i \in \mathbb{F}$. Consider one hidden layer complex-valued neural networks with the {quadratic} activation $\psi$ in the form of
$$ 
\textbf{x} \mapsto \textbf{v}^T \psi(\textbf{Wx})
$$
where $\textbf{W} \in \mathbb{F}^{k\times d}, \textbf{v} \in \mathbb{F}^k$, ${v}_i \neq {0}$, \text{and } $k \geq d$. The training loss as a function of the weights ($\textbf{W}$, $\textbf{v}$) is defined as
\begin{align*}
\mathcal{L}(\textbf{W},\textbf{v}) &= \frac{1}{2n}\sum^n_{i=1}\| y_i - \textbf{v}^T\psi(\textbf{W}\textbf{x}_i)\|^2.
\end{align*}
\end{definition}
Noted that $\| y_i - \textbf{v}^T\psi(\textbf{W}\textbf{x}_i)\|^2 = (y_i - \textbf{v}^T\psi(\textbf{W}\textbf{x}_i))^2$ when $\mathbb{F} = \mathbb{R}$. We prove that one hidden layer CVNNs with {quadratic} activations have no spurious local minima.

\begin{theorem} 
\label{thm_main}
Assume the dataset, loss function, and training model are defined as in Definition \ref{def_main} with $\mathbb{F} = \mathbb{C}$. Then the training loss as a function of the weight $(\text{\normalfont \textbf{W}}$, $\textbf{\normalfont \textbf{v}})$ has no spurious local minima, i.e. all local minima are global.
\end{theorem} 
\begin{proof}
See Appendix \ref{proof_main}.
\end{proof}

Lemma \ref{lem_22} shows that, under the same setting, Theorem \ref{thm_main} does not hold when $\mathbb{F}=\mathbb{R}$, i.e. in real-valued networks poor local minima exist. This provides evidence for a substantial difference between real-valued and complex-valued networks.

\begin{lemma}[Theorem 2, Corollary 3 in \cite{yun2018small}]  
\label{lem_22}
Let the loss function $\mathcal{L}(\textbf{\normalfont \textbf{W}},\textbf{\normalfont \textbf{v}}) $ and the network structure be defined as in Definition \ref{def_main}. Consider the dataset
$$
\textbf{X} = 
\begin{bmatrix}
    \textbf{x}_{1}  & \textbf{x}_{2} & \textbf{x}_{3}
\end{bmatrix}
=
\begin{bmatrix}
    1  & 0 & \frac{1}{2} \\
    0       & 1 & \frac{1}{2}
\end{bmatrix} , 
\textbf{Y} = 
\begin{bmatrix}
    {y}_{1}  & {y}_{2} & {y}_{3}
\end{bmatrix}
=
\begin{bmatrix}
    0  & 0 & 1
\end{bmatrix},
$$
and the weight
$$
\bar{\textbf{\normalfont \textbf{W}} }= 
\begin{bmatrix}
    \bar{\textbf{w} }_{1}   \\
    \bar{\textbf{w} }_{2}
\end{bmatrix}
=
\begin{bmatrix}
    1  & 1 \\
    1       & 1 
 \end{bmatrix}, 
\bar{\textbf{\normalfont \textbf{v}}} = 
\begin{bmatrix}
    \bar{{v} }_{1}   & \bar{{v} }_{2} \\
\end{bmatrix}
=
\begin{bmatrix}
    \frac{1}{6}  & \frac{1}{6}
\end{bmatrix}.
$$
When $\mathbb{F} = \mathbb{R}$ the weight is a poor local minimum of the loss function.
\end{lemma}

\section{Complex-Valued Weights Turn Local Minima into Saddle Points}

In this section, we provide a concrete example of how complex-valued weights turn local minima into saddle points. Let the weight
$$
(\bar{\textbf{\normalfont \textbf{W}} }, \bar{\textbf{\normalfont \textbf{v}}})= 
\left(
\begin{bmatrix}
    1  & 1 \\
    1       & 1 
 \end{bmatrix}, 
\begin{bmatrix}
    \frac{1}{6}  & \frac{1}{6}
\end{bmatrix}\right)
$$
be a local minimum of the real-valued network for the given dataset as defined in the previous section. Now we analyze $\left(\bar{\textbf{W} }, \bar{\textbf{v}}\right)$ in two different networks and in two ways. On one hand, we show that the Hessians at the point are different in real networks and in complex networks. On the other hand, we prove that there is a point with strictly less loss in an arbitrarily small neighborhood of $(\bar{\textbf{W} }, \bar{\textbf{v}})$ in the complex network.

\begin{lemma} 
\label{lem_23}
Assume the dataset, loss function, and training model are defined as above. At $\bar{\textbf{\normalfont \textbf{W}} }$, the Hessian has no negative eigenvalue when $\mathbb{F} = \mathbb{R}$ and has both positive and negative eigenvalues when $\mathbb{F} = \mathbb{C}$.
\end{lemma}
\begin{proof}
See Appendix \ref{proof_lem_23}.
\end{proof}

By analyzing a specific critical point, Lemma \ref{lem_23} shows us that complex-valued weights can turn poor local minima into saddle points. As an alternative way to illustrate the insight, we provide Lemma \ref{lem_24}. The proofs are provided in the appendix.


\begin{lemma} 
\label{lem_24}
Assume the dataset, loss function, and training model are defined as above. When $\mathbb{F} = \mathbb{C}$, within an arbitrarily small neighborhood of $(\bar{\textbf{\normalfont \textbf{W}}},\bar{\textbf{\normalfont \textbf{v}}})$ there is a point $(\hat{\textbf{\normalfont \textbf{W}}},\hat{\textbf{\normalfont \textbf{v}}})$ such that $\mathcal{L}(\hat{\textbf{\normalfont \textbf{W}}},\hat{\textbf{\normalfont \textbf{v}}}) < \mathcal{L}(\bar{\textbf{\normalfont \textbf{W}}},\bar{\textbf{\normalfont \textbf{v}}}) $.
\end{lemma}
\begin{proof}
See Appendix \ref{proof_lem_24}.
\end{proof}

Now, we provide proof sketches for Lemma \ref{lem_23} and Lemma \ref{lem_24}. In the proof of Lemma \ref{lem_23}, we derive the Hessian matrices at $\bar{\textbf{\normalfont \textbf{W}} }$ for real-valued and complex-valued networks and calculate their eigenvalues. We denote the Hessian matrix by $\mathcal{H}^{\mathbb{R}}_{\bar{\textbf{W}}}$ for the real-valued case and $\mathcal{H}^{\mathbb{C}}_{\bar{\textbf{W}}}$ for the complex-valued case. Note that $\mathbb{C}^{n}$ can be treated as $\mathbb{R}^{2n}$ in some sense. Therefore $\mathcal{H}^{\mathbb{C}}_{\bar{\textbf{W}}}$ is four times the size of $\mathcal{H}^{\mathbb{R}}_{\bar{\textbf{W}}}$. By some derivative calculations and substituting the weight and dataset given in Lemma \ref{lem_22} into the expression, we have
$$
\mathcal{H}^{\mathbb{R}}_{\bar{\textbf{W}}} = 
 \begin{bmatrix}
    \frac{7}{108}  & \frac{-1}{108} & \frac{5}{108}  & \frac{1}{108} \\
    \frac{-1}{108}       & \frac{7}{108} & \frac{1}{108}       & \frac{5}{108} \\
    \frac{5}{108}  & \frac{1}{108} & \frac{7}{108}  & \frac{-1}{108}  \\
    \frac{1}{108}       & \frac{5}{108}  & \frac{-1}{108}       & \frac{7}{108} 
\end{bmatrix} 
$$ 
which has no negative eigenvalues, and
$$
\mathcal{H}^{\mathbb{C}}_{\bar{\textbf{W}}} = 
 \begin{bmatrix}
    \mathcal{H}_1  & \mathcal{H}_1&  \mathcal{H}_2  & \mathcal{H}_3 \\
     \mathcal{H}_1     &  \mathcal{H}_1 & \mathcal{H}_3       & \mathcal{H}_2 \\
   \mathcal{H}_2  & \mathcal{H}_3 &  \mathcal{H}_1 &  \mathcal{H}_1  \\
    \mathcal{H}_3      & \mathcal{H}_2  &  \mathcal{H}_1      &  \mathcal{H}_1
\end{bmatrix} 
$$
where 
$$
\mathcal{H}_1 = 
 \begin{bmatrix}
    \frac{5}{216} & \frac{1}{216} \\
    \frac{1}{216}      & \frac{5}{216}
\end{bmatrix}, 
\mathcal{H}_2 = 
 \begin{bmatrix}
    \frac{1}{108} & -\frac{1}{108} \\
     -\frac{1}{108} &  \frac{1}{108}
\end{bmatrix}
$$
and $\mathcal{H}_3$ is a zero matrix.
It can be verified that $\mathcal{H}^{\mathbb{C}}_{\bar{\textbf{W}}}$ has both negative and positive eigenvalues. This illustrates that complex-valued weights turn local minima into saddle points. As an alternative proof, in the proof of Lemma \ref{lem_24}, we make the following permutation on $\bar{\textbf{\normalfont \textbf{W}} }$,
$$
\hat{\textbf{W}} = 
\begin{bmatrix}
    1-\frac{1}{10^N}  & 1+ \frac{i}{10^N} \\
    1-\frac{1}{10^N}  & 1+ \frac{i}{10^N}
\end{bmatrix}
$$ for an arbitrarily large $N\in \mathbb{N}^+$, 
and by simple calculations we can prove that $ \mathcal{L} ( \hat{\textbf{W}},\bar{\textbf{v}}) < \frac{1}{9} = \mathcal{L}(\bar{\textbf{W}},\bar{\textbf{v}})$.

\bibliographystyle{alpha}
\bibliography{biblio}


\newpage
\appendix

\section{Supporting Lemmas}

\begin{lemma} 
\label{lem_41}
If $\widetilde{\textbf{\normalfont \textbf{W}}}$ is a local minimum of $\mathcal{L}(\widetilde{\textbf{\normalfont \textbf{W}}})$, 
\begin{align*}
0 \leq
(\textbf{h}^*,\textbf{h}^T)\cdot  \widetilde{\nabla}^2 \mathcal{L}(\widetilde{\textbf{\normalfont \textbf{W}}}) \cdot
\begin{pmatrix}
\textbf{h} \\
\textbf{h}^C
\end{pmatrix} \in \mathbb{R}
\end{align*}
where
\begin{align*}
\widetilde{\nabla}^2 \mathcal{L}(\widetilde{\textbf{\normalfont \textbf{W}}}) =
\begin{pmatrix}
\nabla_{\textbf{\normalfont \textbf{W}}}\nabla_{\textbf{\normalfont \textbf{W}}^C} \mathcal{L}(\widetilde{\textbf{\normalfont \textbf{W}}}) & \nabla^2_{\textbf{W}^C} \mathcal{L}(\widetilde{\textbf{\normalfont \textbf{W}}})\\
\nabla^2_{\textbf{\normalfont \textbf{W}}} \mathcal{L}(\widetilde{\textbf{\normalfont \textbf{W}}}) & \nabla_{\textbf{\normalfont \textbf{W}}^C}\nabla_{\textbf{\normalfont \textbf{W}}} \mathcal{L}(\widetilde{\textbf{\normalfont \textbf{W}}}) 
\end{pmatrix}
\end{align*}
for all $\textbf{h} \in \mathbb{C}^{kd}$.
\end{lemma}

\begin{proof}
We first prove that it is a real value. By linearity it is sufficient to show
\begin{align*}
(\textbf{h}^*,\textbf{h}^T)\cdot  \widetilde{\nabla}^2 \mathcal{G}_i(\textbf{W}) \cdot
\begin{pmatrix}
\textbf{h} \\
\textbf{h}^C
\end{pmatrix} \in \mathbb{R}.
\end{align*}
Let $\textbf{h} = \text{vec}(\textbf{U})$ be an arbitrary direction. Since  
$$(\nabla^2_{\textbf{W}} \mathcal{G}_i(\textbf{W}))^C = \nabla^2_{\textbf{W}^C} \mathcal{G}_i(\textbf{W})$$ and 
$$(\nabla_{\textbf{W}}\nabla_{\textbf{W}^C} \mathcal{G}_i(\textbf{W}))^C = \nabla_{\textbf{W}^C}\nabla_{\textbf{W}} \mathcal{G}_i(\textbf{W}),$$
we have 
$$(\text{vec}(\textbf{U})^T \nabla_{\textbf{W}^C}\nabla_{\textbf{W}} \mathcal{G}_i(\textbf{W}) \text{vec}(\textbf{U})^C)^C = \text{vec}(\textbf{U})^* \nabla_{\textbf{W}}\nabla_{\textbf{W}^C} \mathcal{G}_i(\textbf{W}) \text{vec}(\textbf{U})$$ and 
$$(\text{vec}(\textbf{U})^T \nabla^2_{\textbf{W}} \mathcal{G}_i(\textbf{W}) \text{vec}(\textbf{U}))^C = \text{vec}(\textbf{U})^* \nabla^2_{\textbf{W}^C} \mathcal{G}_i(\textbf{W}) \text{vec}(\textbf{U}^C).$$
Thus,
\begin{align*}
&(\text{vec}(\textbf{U})^*, \text{vec}(\textbf{U})^T) 
\begin{pmatrix}
\nabla_{\textbf{W}}\nabla_{\textbf{W}^C} \mathcal{G}_i(\textbf{W}) & \nabla^2_{\textbf{W}^C} \mathcal{G}_i(\textbf{W})\\
\nabla^2_{\textbf{W}} \mathcal{G}_i(\textbf{W}) & \nabla_{\textbf{W}^C}\nabla_{\textbf{W}} \mathcal{G}_i(\textbf{W})  
\end{pmatrix}
\begin{pmatrix}
\text{vec}(\textbf{U}) \\
\text{vec}(\textbf{U})^C
\end{pmatrix}\\
&=( \text{vec}(\textbf{U})^T \nabla^2_{\textbf{W}} \mathcal{G}_i(\textbf{W}) \text{vec}(\textbf{U}) + \text{vec}(\textbf{U})^T \nabla_{\textbf{W}^C}\nabla_{\textbf{W}} \mathcal{G}_i(\textbf{W}) \text{vec}(\textbf{U}^C)\\ 
&+ \text{vec}(\textbf{U})^* \nabla_{\textbf{W}}\nabla_{\textbf{W}^C} \mathcal{G}_i(\textbf{W}) \text{vec}(\textbf{U}) + \text{vec}(\textbf{U})^* \nabla^2_{\textbf{W}^C} \mathcal{G}_i(\textbf{W}) \text{vec}(\textbf{U}^C))\\
&= 2\mathcal{R}(\text{vec}(\textbf{U})^T \nabla^2_{\textbf{W}} \mathcal{G}_i(\textbf{W}) \text{vec}(\textbf{U}) +\text{vec}(\textbf{U})^* \nabla_{\textbf{W}}\nabla_{\textbf{W}^C} \mathcal{G}_i(\textbf{W}) \text{vec}(\textbf{U})) \in \mathbb{R} \\
\end{align*}
Now suppose $\widetilde{\textbf{W}}$ is a local minimum,  the second order expansion at $\widetilde{\textbf{W}}$ is
\begin{align*}
\mathcal{L}(\widetilde{\textbf{W}} + \textbf{U})= \mathcal{L}(\widetilde{\textbf{W}}) +  \widetilde{\nabla} \mathcal{L}(\widetilde{\textbf{W}}) \cdot 
\begin{pmatrix}
\textbf{h} \\
\textbf{h}
\end{pmatrix} +
\frac{1}{2} (\textbf{h},\textbf{h}^*)\cdot   \widetilde{\nabla}^2 \mathcal{L}(\widetilde{\textbf{W}}) \cdot
\begin{pmatrix}
\textbf{h} \\
\textbf{h}^*
\end{pmatrix} + o(||\textbf{h}||^2).
\end{align*}
Since $\widetilde{\textbf{W}}$ is a local minimum, the gradient is zero. As in the standard proof, when $||\textbf{h}||$ is small enough,
$$
\frac{1}{2} (\textbf{h},\textbf{h}^*)\cdot   \widetilde{\nabla}^2 \mathcal{L}(\widetilde{\textbf{W}}) \cdot
\begin{pmatrix}
\textbf{h} \\
\textbf{h}^*
\end{pmatrix} = 
\mathcal{L}(\widetilde{\textbf{W}} + \textbf{U}) - \mathcal{L}(\widetilde{\textbf{W}}) \geq 0.
$$
\end{proof}

\begin{lemma} 
\label{lem_42}
Any point $\widetilde{\textbf{\normalfont \textbf{W}}} \in \mathbb{C}^{k \times d}$ obeying
$$
\frac{1}{2n} \sum^n_{i=1}(\textbf{x}^T_i \widetilde{\textbf{\normalfont \textbf{W}}}^T \text{\normalfont diag(\textbf{v})} \widetilde{\textbf{\normalfont \textbf{W}}}\textbf{x}_i - y_i)^*\textbf{x}_i\textbf{x}_i^T =\frac{1}{2n}\sum_{i=1}^n\mathcal{L}_i^*(\widetilde{\textbf{\normalfont \textbf{W}}})\textbf{x}_i\textbf{x}_i^T= 0
$$
is a global optimum of the loss function
\begin{align*}
\mathcal{L}(\textbf{\normalfont \textbf{W}}) &= \frac{1}{2n}\sum^n_{i=1}\parallel \textbf{x}^T_i \textbf{\normalfont \textbf{W}}^T \text{\normalfont diag(\textbf{v})} \textbf{\normalfont \textbf{W}}\textbf{x}_i - y_i\parallel^2\\
&= \frac{1}{2n}\sum^n_{i=1}\parallel y_i - \textbf{\normalfont \textbf{v}}^T\psi(\textbf{\normalfont \textbf{W}}\textbf{x}_i)\parallel^2.
\end{align*}
\end{lemma}

\begin{proof}
Let $\textbf{M} = \textbf{W}^T \text{\normalfont diag(\textbf{v})} \textbf{W}$. Then loss function becomes $\mathcal{L}(\textbf{M}) = \frac{1}{2n}\sum^n_{i=1}\parallel \textbf{x}_i^T \textbf{M}\textbf{x}_i - y_i\parallel^2 $.
By some algebra, we write
$$
\mathcal{L}(\textbf{M}) = \frac{1}{2n}\sum^n_{i=1} (\textbf{x}^*_i\textbf{M}^*\textbf{x}_i^{T*}\textbf{x}^T_i\textbf{Mx}_i - 2\mathcal{R}(y_i^*\textbf{x}_i^T\textbf{Mx}_i) + \parallel y_i \parallel^2).
$$
Notice that $\textbf{x}^*_i\textbf{M}^*\textbf{x}_i^{T*}\textbf{x}^T_i\textbf{Mx}_i =  \parallel \textbf{x}^T_i\textbf{Mx}_i \parallel^2$. From the expression we can see $\mathcal{L}(\textbf{M})$ is convex in $\textbf{M}$ because $\mathcal{R}(y_i^*\textbf{x}_i^T\textbf{Mx}_i)$ and $\textbf{x}^T_i\textbf{Mx}_i$ are linear with respect to $\textbf{M}$. 
Now by Wirtinger calculus and the convexity,
$$
\widetilde{\textbf{M}} \text{ being a global minimum of } \mathcal{L}(\textbf{M}) \Leftrightarrow \frac{1}{2n} \sum^n_{i=1}(\textbf{x}^T_i \widetilde{\textbf{M}}\textbf{x}_i - y_i)^* \textbf{x}_i\textbf{x}_i^T = 0.
$$
Note that $\widetilde{\textbf{M}} = \widetilde{\textbf{W}}^T \text{\normalfont diag(\textbf{v})} \widetilde{\textbf{W}}$ for some $\widetilde{\textbf{W}}$.
Thus, for any arbitrary $\textbf{M}\in \mathbb{C}^{d\times d}$ we have
$$
\mathcal{L}(\textbf{M}) \geq \mathcal{L}(\widetilde{\textbf{M}})
$$
which implies that for any $\textbf{W} \in \mathbb{C}^{k \times d}$
$$
\mathcal{L}(\textbf{W}) \geq \mathcal{L}(\widetilde{\textbf{W}}).
$$
\end{proof}

\section{Proof of Theorem \ref{thm_main}}
\label{proof_main}

\paragraph{Step 1: Derivative calculations.} First, we demonstrate how to calculate the derivative of $\mathcal{L}(\textbf{W})$. First, we observe that $\mathcal{L}(\textbf{W})$ is a function which maps complex input to real output, i.e. $\mathcal{L}(\textbf{W}): \mathbb{C}^{k\times d} \mapsto \mathbb{R}$, and it is not differentiable because conjugate functions do not satisfy the Cauchy-Riemann Equation. By letting $\mathcal{L}(\textbf{W}) = \frac{1}{2n} \sum^n_{i=1} \mathcal{L}_i(\textbf{W})^*\mathcal{L}_i(\textbf{W})$ where  $\mathcal{L}_i(\textbf{W}): \mathbb{C}^{k\times d} \mapsto \mathbb{C}$ given by $\mathcal{L}_i(\textbf{W}) = \textbf{v}^T\psi(\textbf{W}\textbf{x}_i) - y_i$, $\mathcal{L}_i(\textbf{W})$ is complex differentiable in the traditional sense. Thus, $\mathcal{L}_i(\textbf{W})$ has well-defined first and second derivatives. For a fixed $i$, we let $\mathcal{G}_i(\textbf{W}) = \mathcal{L}_i(\textbf{W})^*\mathcal{L}_i(\textbf{W})$. Now, we show how to calculate the derivatives of $\mathcal{L}_i(\textbf{W})$ and $\mathcal{G}_i(\textbf{W})$. The derivatives of $\mathcal{L}(\textbf{W})$ follow easily by linearity. Denote $\nabla_{\textbf{w}_q}\mathcal{L}_i(\textbf{W})$ the first derivative with respect to the $q$-th row of $\textbf{W}$, and we derive
$$
\nabla_{\textbf{w}_q}\mathcal{L}_i(\textbf{W}) = v_q \psi^{'}(\langle \textbf{w}_q, \textbf{x}_i \rangle)\textbf{x}_i,
$$
$$
\nabla_{\textbf{W}}\mathcal{L}_i(\textbf{W}) = \textbf{D}_{\textbf{v}} \psi^{'}( \textbf{W} \textbf{x}_i )\textbf{x}_i^T,
$$
where $ \textbf{D}_{\textbf{v}} =$ diag$(v_1,\dots,v_k)$. The second derivative can be expressed as
$$
\frac{\partial^2}{\partial\textbf{w}^2_p} \mathcal{L}_i(\textbf{W}) = v_p \psi^{''}(\langle \textbf{w}_p, \textbf{x}_i \rangle)\textbf{x}_i\textbf{x}_i^T,
$$
and
$$
\frac{\partial^2}{\partial\textbf{w}_p\textbf{w}_q} \mathcal{L}_i(\textbf{W}) = 0,
$$
for $p \neq q$. Next, we derive the derivative of $\mathcal{G}_i(\textbf{W})$. By the product rule of Wirtinger calculus, we have
$$
\nabla_{\textbf{W}} \mathcal{G}_i(\textbf{W}) =\nabla_{\textbf{W}} (\mathcal{L}_i^* \mathcal{L}_i)(\textbf{W})  = \nabla_{\textbf{W}} \mathcal{L}_i^*(\textbf{W}) \mathcal{L}_i(\textbf{W}) + \nabla_{\textbf{W}} \mathcal{L}_i(\textbf{W}) \mathcal{L}_i^*(\textbf{W}) = \nabla_{\textbf{W}} \mathcal{L}_i(\textbf{W}) \mathcal{L}_i^*(\textbf{W}).
$$
Note that $ \nabla_{\textbf{W}} \mathcal{L}_i^*(\textbf{W}) \mathcal{L}_i(\textbf{W}) = 0$ since $\mathcal{L}_i^*$ is conjugate-complex differentiable. Similarly,
$$
\nabla_{\textbf{W}^C} \mathcal{G}_i(\textbf{W}) =\nabla_{\textbf{W}^C} (\mathcal{L}_i^* \mathcal{L}_i)(\textbf{W})  = \nabla_{\textbf{W}^C} \mathcal{L}_i^*(\textbf{W}) \mathcal{L}_i(\textbf{W}) + \nabla_{\textbf{W}^C} \mathcal{L}_i(\textbf{W}) \mathcal{L}_i^*(\textbf{W}) = \nabla_{\textbf{W}^C} \mathcal{L}_i^*(\textbf{W}) \mathcal{L}_i(\textbf{W}).
$$
Based on the above equalities, we obtain the second derivatives
\begin{align*}
\nabla^2_{\textbf{W}} \mathcal{G}_i(\textbf{W}) &= \nabla^2_{\textbf{W}} \mathcal{L}_i(\textbf{W}) \mathcal{L}_i^*(\textbf{W}),\\
\nabla^2_{\textbf{W}^C} \mathcal{G}_i(\textbf{W}) &= \nabla^2_{\textbf{W}^C} \mathcal{L}_i^*(\textbf{W}) \mathcal{L}_i(\textbf{W}),\\
\nabla_{\textbf{w}_p}\nabla_{\textbf{w}_p^C} \mathcal{G}_i(\textbf{W}) &= v_p^*v_p \psi^{'}(\langle \textbf{w}_p, \textbf{x}_i \rangle)^*\psi^{'}(\langle \textbf{w}_p, \textbf{x}_i \rangle) \textbf{x}_i^C \textbf{x}_i^T,\\
\nabla_{\textbf{w}_q}\nabla_{\textbf{w}_p^C} \mathcal{G}_i(\textbf{W}) &= v_p^*v_q \psi^{'}(\langle \textbf{w}_p, \textbf{x}_i \rangle)^*\psi^{'}(\langle \textbf{w}_q, \textbf{x}_i \rangle) \textbf{x}_i^C \textbf{x}_i^T, \\
\nabla_{\textbf{w}_p^C}\nabla_{\textbf{w}_p} \mathcal{G}_i(\textbf{W}) &= v_p^*v_p \psi^{'}(\langle \textbf{w}_p, \textbf{x}_i \rangle)^*\psi^{'}(\langle \textbf{w}_p, \textbf{x}_i \rangle) \textbf{x}_i \textbf{x}_i^*,\\
 \nabla_{\textbf{w}_q^C}\nabla_{\textbf{w}_p} \mathcal{G}_i(\textbf{W}) &= v_pv_q^* \psi^{'}(\langle \textbf{w}_p, \textbf{x}_i \rangle)\psi^{'}(\langle \textbf{w}_q, \textbf{x}_i \rangle)^* \textbf{x}_i \textbf{x}_i^*.
\end{align*}
Notice that $\nabla_{\textbf{W}}\nabla_{\textbf{W}^C} \mathcal{G}_i(\textbf{W})$ and $\nabla_{\textbf{W}^C}\nabla_{\textbf{W}} \mathcal{G}_i(\textbf{W})$ are $kd \times kd$ matrices.

\paragraph{Step 2: $\textbf{W}$ being a local minimum implies ($\textbf{W}$, $\textbf{v}$) being a local minimum.} 
We have both weights optimized, i.e. under the condition of both $\textbf{W}$ and $\textbf{v}$ being local minima. However, it is sufficient to utilize only the fact that $\textbf{W}$ is a local minimum. In this proof $\textbf{v}$ can be any vectors with non-zero entries including the local minima. A key observation is that given $v_i \neq 0$, a global minimum of $\mathcal{L}$ w.r.t $\textbf{W}$ is a global minimum w.r.t. ($\textbf{W}$, $\textbf{v}$). The reason is that for any current $v_i$ and the \enquote{targeted} $v_i^{*}$, $\mathcal{L}$ has the same loss with either $(\textbf{w}_i, v_i^{*})$ or $\left(\sqrt{\frac{v_i^{*}}{v_i}}\cdot\textbf{w}_i, v_i\right)$ on the $i^{th}$ row. Consider a toy example, $\psi(\sqrt{2}i \cdot [1,2])\cdot 3 =\psi([1,2]) \cdot -6$, where $v_i = 3$, $v_i^{*}=-6$, and $\sqrt{\frac{v_i^{*}}{v_i}} = \sqrt{2}i$. Therefore, proving any local minimum of $\mathcal{L}$ w.r.t $\textbf{W}$ is a global minimum is sufficient to prove Theorem \ref{thm_main}.

\paragraph{Step 3: Simplifying $\mathcal{H}$.} Let $\textbf{W} \in \mathbb{C}^{k \times d} $ be a local minimum. Let $\textbf{U} \in \mathbb{C}^{k \times d} $ be an arbitrary direction and $\textbf{h} = \text{vec}(\textbf{U})$.
We define
\begin{align*}
\mathcal{H} &= \frac{1}{2} (\textbf{h}^*,\textbf{h}^T)\cdot   \widetilde{\nabla}^2 \mathcal{L}(\textbf{W}) \cdot
\begin{pmatrix}
\textbf{h} \\
\textbf{h}^C
\end{pmatrix}  \\&= \frac{1}{2} (\text{vec}(\textbf{U})^*, \text{vec}(\textbf{U})^T) 
\begin{pmatrix}
\nabla_{\textbf{W}}\nabla_{\textbf{W}^C} \mathcal{L}(\textbf{W}) & \nabla^2_{\textbf{W}^C} \mathcal{L}(\textbf{W})\\
\nabla^2_{\textbf{W}} \mathcal{L}(\textbf{W}) & \nabla_{\textbf{W}^C}\nabla_{\textbf{W}} \mathcal{L}(\textbf{W}) 
\end{pmatrix}
\begin{pmatrix}
\text{vec}(\textbf{U}) \\
\text{vec}(\textbf{U})^C
\end{pmatrix}
\end{align*}
By linearity,
\begin{align*}
\mathcal{H} &= \frac{1}{2} (\text{vec}(\textbf{U})^*, \text{vec}(\textbf{U})^T) 
\begin{pmatrix}
\frac{1}{2n}\sum^n_{i=1}\nabla_{\textbf{W}}\nabla_{\textbf{W}^C} \mathcal{G}_i(\textbf{W}) &\frac{1}{2n}\sum^n_{i=1} \nabla^2_{\textbf{W}^C} \mathcal{G}_i(\textbf{W})\\
\frac{1}{2n}\sum^n_{i=1}\nabla^2_{\textbf{W}} \mathcal{G}_i(\textbf{W}) & \frac{1}{2n}\sum^n_{i=1}\nabla_{\textbf{W}^C}\nabla_{\textbf{W}} \mathcal{G}_i(\textbf{W})
\end{pmatrix}
\begin{pmatrix}
\text{vec}(\textbf{U}) \\
\text{vec}(\textbf{U})^C
\end{pmatrix}.
\end{align*}
For each term we have
\begin{align*}
(\text{vec}(\textbf{U})^*, \text{vec}(\textbf{U})^T) 
\begin{pmatrix}
\nabla_{\textbf{W}}\nabla_{\textbf{W}^C} \mathcal{G}_i(\textbf{W}) & \nabla^2_{\textbf{W}^C} \mathcal{G}_i(\textbf{W}) \\
\nabla^2_{\textbf{W}} \mathcal{G}_i(\textbf{W}) & \nabla_{\textbf{W}^C}\nabla_{\textbf{W}} \mathcal{G}_i(\textbf{W}) 
\end{pmatrix}
\begin{pmatrix}
\text{vec}(\textbf{U}) \\
\text{vec}(\textbf{U})^C
\end{pmatrix}\\
= 2\mathcal{R}(\text{vec}(\textbf{U})^T \nabla^2_{\textbf{W}} \mathcal{G}_i(\textbf{W}) \text{vec}(\textbf{U}) +\text{vec}(\textbf{U})^* \nabla_{\textbf{W}}\nabla_{\textbf{W}^C} \mathcal{G}_i(\textbf{W}) \text{vec}(\textbf{U})).
\end{align*}
Now we consider two cases. Case 1 is for $\text{rank}(\textbf{D}_{\textbf{v}}\textbf{W}) = d$ and case 2 is for $\text{rank}(\textbf{D}_{\textbf{v}}\textbf{W}) < d$.  
For the first case, since $k \geq d$ and $\text{rank}(\textbf{D}_{\textbf{v}}\textbf{W}) = d$, $\textbf{D}_{\textbf{v}}\textbf{W} $ has a left inverse $\textbf{K} \in \mathbb{C}^{d\times k}$ such that $\textbf{K} \textbf{D}_{\textbf{v}}\textbf{W}  = \textbf{I}$. Notice that by $\nabla_{\textbf{W}} \mathcal{L}(\textbf{W}) =0$ and $\nabla_{\textbf{W}} \mathcal{L}(\textbf{W}) = \frac{1}{2n} \sum_{i=1}^n \nabla_{\textbf{W}} \mathcal{L}_i(\textbf{W})\mathcal{L}^*_i(\textbf{W})  =  \textbf{D}_{\textbf{v}} \textbf{W} (\frac{1}{n} \sum_{i=1}^n \mathcal{L}^*_i(\textbf{W}) \textbf{x}_i \textbf{x}_i^T)$, we have
$$
\textbf{D}_{\textbf{v}} \textbf{W} (\frac{1}{n} \sum_{i=1}^n \mathcal{L}^*_i(\textbf{W}) \textbf{x}_i \textbf{x}_i^T) = 0.
$$
Multiplying both sides by $\textbf{K}$ we get
$$
\frac{1}{n} \sum_{i=1}^n \mathcal{L}^*_i(\textbf{W}) \textbf{x}_i \textbf{x}_i^T = 0,
$$
which concludes the proof by Lemma \ref{lem_42}.
For the second case, we can let $\textbf{U}=\textbf{a}\textbf{b}^T$ with $\textbf{a}\in\mathbb{C}^k$  and $\textbf{D}_{\textbf{v}} \textbf{a}\in \text{Null}(\textbf{W}^T)$. $\textbf{b}\in \mathbb{C}^d$ is an arbitrary vector. We now show that $\text{vec}(\textbf{U})^* \nabla_{\textbf{W}}\nabla_{\textbf{W}^C} \mathcal{G}_i(\textbf{W}) \text{vec}(\textbf{U}) = 0$. 
Recall that
$$ \nabla_{\textbf{w}_p^C} \mathcal{G}_i(\textbf{W}) = \nabla_{\textbf{w}_p^C} \mathcal{L}_i^*(\textbf{W})\mathcal{L}_i(\textbf{W}) = v_p^* \psi^{'}(\langle \textbf{w}_p, \textbf{x}_i \rangle)^* \textbf{x}_i^C \mathcal{L}_i(\textbf{W})$$ 
$$ \nabla_{\textbf{w}_p}\nabla_{\textbf{w}_p^C} \mathcal{G}_i(\textbf{W}) = v_p^*v_p \psi^{'}(\langle \textbf{w}_p, \textbf{x}_i \rangle)^*\psi^{'}(\langle \textbf{w}_p, \textbf{x}_i \rangle) \textbf{x}_i^C \textbf{x}_i^T = \parallel v_p \psi^{'}(\langle \textbf{w}_p, \textbf{x}_i \rangle) \parallel^2 \textbf{x}_i^C \textbf{x}_i^T  $$
$$ \nabla_{\textbf{w}_q}\nabla_{\textbf{w}_p^C} \mathcal{G}_i(\textbf{W}) = v_p^*v_q \psi^{'}(\langle \textbf{w}_p, \textbf{x}_i \rangle)^*\psi^{'}(\langle \textbf{w}_q, \textbf{x}_i \rangle) \textbf{x}_i^C \textbf{x}_i^T. $$
Therefore we can treat $\nabla_{\textbf{W}}\nabla_{\textbf{W}^C} \mathcal{G}_i(\textbf{W}) $ as a $k \times k$ matrix with each entry being a $d \times d$ matrix.
Now by some algebra we have
$$
\text{vec}(\textbf{U})^* \nabla_{\textbf{W}}\nabla_{\textbf{W}^C} \mathcal{G}_i(\textbf{W}) \text{vec}(\textbf{U}) = \parallel \textbf{x}_i^T \textbf{W}^T \textbf{D}_{\textbf{v}} \textbf{U} \textbf{x}_i \parallel^2
$$
where
$$
\textbf{x}_i^T \textbf{W}^T \textbf{D}_{\textbf{v}} \textbf{U} \textbf{x}_i = \textbf{x}_i^T \textbf{W}^T \textbf{D}_{\textbf{v}} \textbf{a} \textbf{b}^T \textbf{x}_i = 0.
$$
Now by linearity and Lemma \ref{lem_41} we have
\begin{align*}
\mathcal{H} &=  \frac{1}{2n}\mathcal{R}(\text{vec}(\textbf{U})^T \sum^n_{i=1}\nabla^2_{\textbf{W}} \mathcal{G}_i(\textbf{W}) \text{vec}(\textbf{U})) \geq 0
\end{align*}
where
\begin{align*}
\sum_{i=1}^n\text{vec}(\textbf{U})^T \nabla^2_{\textbf{W}} \mathcal{G}_i(\textbf{W}) \text{vec}(\textbf{U}) 
&= \sum_{i=1}^n\text{vec}(\textbf{U})^T \nabla^2_{\textbf{W}} \mathcal{L}_i(\textbf{W}) \mathcal{L}_i^*(\textbf{W}) \text{vec}(\textbf{U})\\
&= 2\sum_{i=1}^n\mathcal{L}_i^*(\textbf{W})(\textbf{x}_i^T\textbf{U}^T\textbf{D}_{\textbf{v}}\textbf{U}\textbf{x}_i)\\
&= 2(\textbf{a}^T\textbf{D}_{\textbf{v}}\textbf{a})\textbf{b}^T\left(\sum_{i=1}^n\mathcal{L}_i^*(\textbf{W})\textbf{x}_i\textbf{x}_i^T\right)\textbf{b}.
\end{align*}
We argue that we can assume $(\textbf{a}^T\textbf{D}_{\textbf{v}}\textbf{a}) \neq 0$ here. The reason is the following. Since $\textbf{a} \neq \textbf{0}$, there is an entry $a_i \neq 0$. Suppose $\textbf{a}^T\textbf{D}_{\textbf{v}}\textbf{a} = 0$, we can multiply $v_i$ by $\frac{1}{4}$ and multiply the the $i'$th row of $\textbf{W}$ by $2$. Now $\textbf{a}^T\textbf{D}_{\textbf{v}_{new}}\textbf{a} = -\frac{3}{4}v_ia_i^2 \neq 0$. Note that the two weight matrices $\textbf{W}$ and $\textbf{W}_{new}$ have the same null space and $\textbf{W}_{new}$ is also a local minimum. By Lemma \ref{lem_42} the old matrix $\textbf{W}$ together with $\textbf{v}$ is a global minimum if and only if the new matrix $\textbf{W}_{new}$ together with $\textbf{v}_{new}$ is a global minimum, because their corresponding $\textbf{M} = \textbf{W}^T\text{diag}(\textbf{v})\textbf{W} $ is the same. Therefore, proving $\textbf{W}_{new}$ is a global minimum of $\mathcal{L}$ is equivalent to proving $\textbf{W}$ is a global minimum. Thus, we can assume $(\textbf{a}^T\textbf{D}_{\textbf{v}}\textbf{a}) \neq 0$ without loss of generality.

\paragraph{Step 4: Proving $\sum_{i=1}^n\mathcal{L}_i^*(\textbf{W})\textbf{x}_i\textbf{x}_i^T = 0$.} 
Let $2(\textbf{a}^T\textbf{D}_{\textbf{v}}\textbf{a}) = a_1 + ia_2 \in \mathbb{C}$ and  $\textbf{b}^T\left(\sum_{i=1}^n\mathcal{L}_i^*(\textbf{W})\textbf{x}_i\textbf{x}_i^T\right)\textbf{b} = b_1 + ib_2 \in \mathbb{C}$. We now prove $\sum_{i=1}^n\mathcal{L}_i^*(\textbf{W})\textbf{x}_i\textbf{x}_i^T = 0$ by contradiction. Since $\mathcal{H} = \frac{1}{2n}\mathcal{R}((a_1 + a_2i)\cdot(b_1 + b_2i)) = \frac{1}{2n}(a_1b_1 - a_2b_2) \geq 0$, we prove if $\sum_{i=1}^n\mathcal{L}_i^*(\textbf{W})\textbf{x}_i\textbf{x}_i^T \neq 0$ then $\mathcal{H} < 0$ for some $(b_1, b_2)$. Since for a fixed pair $(a_1,a_2)$ we can make $a_1b_1 - a_2b_2$ a negative number simply by setting the signs of $(b_1, b_2)$ according to the signs of $(a_1,a_2)$. For example, if $a_1 >0$ and $a_2 <0$ then $a_1b_1 - a_2b_2<0$ for $(b_1, b_2)$ with $b_1<0$ and $b_2<0$. Now let
$$
\mathcal{M} = \sum_{i=1}^n\mathcal{L}_i^*(\textbf{W})\textbf{x}_i\textbf{x}_i^T \neq 0,  \mathcal{M} \in \mathbb{C}^{d\times d}.
$$
Let $\mathcal{M}_{i,j}$ denotes the entry on the $i$'th row and the $j$'th column of  $\mathcal{M}$. Now, we show that we can have any sign on $b_1$ and $b_2$, which implies $\mathcal{M}$ must be zero. Suppose there exists a $i \in [d]$ such that $\mathcal{M}_{i,i} \neq 0$, then we let $\textbf{b} = (0,\dots,\beta_i,\dots,0)^T$ where $\beta_i$ can be any complex number. Therefore, $b_1+ib_2 = \mathcal{M}_{i,i} \cdot \beta_i^2$ can have any sign. Suppose $\mathcal{M}_{i,i} = 0$ for all $i \in [d]$ and $\mathcal{M}_{i,j} \neq 0$ for some $(i,j)$, then we let $\textbf{b} = (0,\dots,\beta_i,\dots,0,\dots,\beta_j,\dots,0)^T$. Now $b_1+ib_2 = 2 \mathcal{M}_{i,j}\cdot \beta_i\cdot\beta_j$ which can have any sign. Thus, if $\sum_{i=1}^n\mathcal{L}_i^*(\textbf{W})\textbf{x}_i\textbf{x}_i^T \neq 0$, then $\mathcal{H} < 0$ for some $(b_1, b_2)$.
Therefore, $\textbf{W}$ is a global minimum by Lemma \ref{lem_42}, concluding the proof.

\section{Proof of Lemma \ref{lem_23}}
\label{proof_lem_23}
Recall that
{\it
\begin{align*}
\mathcal{L}(\textbf{\normalfont \textbf{W}},\textbf{\normalfont \textbf{v}}) &= \frac{1}{2n}\sum^n_{i=1}\parallel y_i - \textbf{\normalfont \textbf{v}}^T\psi(\textbf{\normalfont \textbf{W}}\textbf{x}_i)\parallel^2,
\end{align*}
$$
\textbf{X} = 
\begin{bmatrix}
    \textbf{x}_{1}  & \textbf{x}_{2} & \textbf{x}_{3}
\end{bmatrix}
=
\begin{bmatrix}
    1  & 0 & \frac{1}{2} \\
    0       & 1 & \frac{1}{2}
\end{bmatrix} , 
$$
$$
\textbf{Y} = 
\begin{bmatrix}
    {y}_{1}  & {y}_{2} & {y}_{3}
\end{bmatrix}
=
\begin{bmatrix}
    0  & 0 & 1
\end{bmatrix},
$$
$$
\bar{\textbf{\normalfont \textbf{W}} }= 
\begin{bmatrix}
    \bar{\textbf{w} }_{1}   \\
    \bar{\textbf{w} }_{2}
\end{bmatrix}
=
\begin{bmatrix}
    1  & 1 \\
    1       & 1 
 \end{bmatrix}, 
 $$
 $$
\bar{\textbf{\normalfont \textbf{v}}} = 
\begin{bmatrix}
    \bar{{v} }_{1}   & \bar{{v} }_{2} \\
\end{bmatrix}
=
\begin{bmatrix}
    \frac{1}{6}  & \frac{1}{6}
\end{bmatrix}.
$$
}
It is to be noticed that the Hessian at $\bar{\textbf{W} }$ is
$$
\mathcal{H}^{\mathbb{R}}_{\bar{\textbf{W}}}= 
\begin{bmatrix}
    \frac{\partial^2}{\partial \bar{\textbf{w} }_{1}^2} \mathcal{L}(\textbf{W},\textbf{v})  & \frac{\partial^2}{\partial \bar{\textbf{w} }_{1}\bar{\textbf{w} }_{2}} \mathcal{L}(\textbf{W},\textbf{v}) \\
    \frac{\partial^2}{\partial \bar{\textbf{w} }_{2}\bar{\textbf{w} }_{1}} \mathcal{L}(\textbf{W},\textbf{v})       & \frac{\partial^2}{\partial \bar{\textbf{w} }_{2}^2} \mathcal{L}(\textbf{W},\textbf{v})
 \end{bmatrix}
 $$
where
\begin{align*}
 \frac{\partial^2}{\partial \bar{\textbf{w} }_{1}^2} \mathcal{L}({\textbf{W}},\textbf{v}) &= \frac{\bar{v}_1}{n} \sum^n_{i=1}(\bar{\textbf{v}}^T\psi(\bar{\textbf{W}}\textbf{x}_i)- y_i )\psi^{''}(\bar{\textbf{w} }_{1} \textbf{x}_{i}) \textbf{x}_{i}  \textbf{x}_{i}^T + \frac{\bar{v}_1^2}{n}  \sum^n_{i=1}(\psi^{'}(\bar{\textbf{w} }_{1} \textbf{x}_{i}))^2 \textbf{x}_{i}  \textbf{x}_{i}^T \\
 & = 
 \begin{bmatrix}
    \frac{7}{108}  & \frac{-1}{108} \\
    \frac{-1}{108}       & \frac{7}{108}
\end{bmatrix} \\
&= \frac{\bar{v}_2}{n} \sum^n_{i=1}( \bar{\textbf{v}}^T\psi(\bar{\textbf{W}}\textbf{x}_i)-y_i )\psi^{''}(\bar{\textbf{w} }_{2} \textbf{x}_{i}) \textbf{x}_{i}  \textbf{x}_{i}^T + \frac{\bar{v}_2^2}{n}  \sum^n_{i=1}(\psi^{'}(\bar{\textbf{w} }_{2} \textbf{x}_{i}))^2 \textbf{x}_{i}  \textbf{x}_{i}^T \\
 & = 
 \frac{\partial^2}{\partial \bar{\textbf{w} }_{2}^2} \mathcal{L}({\textbf{W}},\textbf{v}) ,
\end{align*}
and
\begin{align*}
\frac{\partial^2}{\partial \bar{\textbf{w} }_{1}\bar{\textbf{w} }_{2}} \mathcal{L}(\textbf{W},\textbf{v}) &=  \frac{\bar{v}_1 \bar{v}_2}{n}  \sum^n_{i=1}\psi^{'}(\bar{\textbf{w} }_{1} \textbf{x}_{i})\psi^{'}(\bar{\textbf{w} }_{2} \textbf{x}_{i}) \textbf{x}_{i}  \textbf{x}_{i}^T \\
 & = 
 \begin{bmatrix}
    \frac{5}{108}  & \frac{1}{108} \\
    \frac{1}{108}       & \frac{5}{108}
\end{bmatrix} \\
&= \frac{\bar{v}_2 \bar{v}_1}{n}  \sum^n_{i=1}\psi^{'}(\bar{\textbf{w} }_{2} \textbf{x}_{i})\psi^{'}(\bar{\textbf{w} }_{1} \textbf{x}_{i}) \textbf{x}_{i}  \textbf{x}_{i}^T \\
 & = 
 \frac{\partial^2}{\partial \bar{\textbf{w} }_{2}\bar{\textbf{w} }_{1}} \mathcal{L}(\textbf{W},\textbf{v}).
\end{align*}
It can be verified easily that $ \mathcal{H}^{\mathbb{R}}_{\textbf{W}} $ has no negative eigenvalue. Now we analyze the Wirtinger Hessians at  $\bar{\textbf{W} }$, namely the Hessian at the complex-valued setting. Recall that 
\begin{align*}
\mathcal{H}^{\mathbb{C}}_{\bar{\textbf{W}}} =
\begin{pmatrix}
\nabla_{\bar{\textbf{W}}}\nabla_{\bar{\textbf{W}}^C} \mathcal{L}({{\textbf{W}}},\textbf{v}) & \nabla^2_{\bar{\textbf{W}}^C} \mathcal{L}({\textbf{W}},\textbf{v})\\
\nabla^2_{\bar{\textbf{W}}} \mathcal{L}({{\textbf{W}}},\textbf{v}) & \nabla_{\bar{\textbf{W}}^C}\nabla_{\bar{\textbf{W}}} \mathcal{L}({\textbf{W}},\textbf{v}) 
\end{pmatrix}
\end{align*}
and we calculate it term by term. Firstly we notice that $\nabla_{\bar{\textbf{W}}}\nabla_{\bar{\textbf{W}}^C} \mathcal{L}({{\textbf{W}}},\textbf{v}) = \nabla_{\bar{\textbf{W}}^C}\nabla_{\bar{\textbf{W}}} \mathcal{L}({\textbf{W}},\textbf{v}) $ and $  \nabla^2_{\bar{\textbf{W}}^C} \mathcal{L}({\textbf{W}},\textbf{v}) =  \nabla^2_{\bar{\textbf{W}}} \mathcal{L}({{\textbf{W}}},\textbf{v}) $ because the weights and data are real-valued. Now we have
$$
\nabla_{\bar{\textbf{W}}}\nabla_{\bar{\textbf{W}}^C} \mathcal{L}({{\textbf{W}}}) = 
 \begin{bmatrix}
    \nabla_{\bar{\textbf{w}}_1}\nabla_{\bar{\textbf{w}}^C_1} \mathcal{L}({{\textbf{W}}},\textbf{v})  & \nabla_{\bar{\textbf{w}}_1}\nabla_{\bar{\textbf{w}}^C_2} \mathcal{L}({{\textbf{W}}},\textbf{v}) \\
    \nabla_{\bar{\textbf{w}}_2}\nabla_{\bar{\textbf{w}}^C_1} \mathcal{L}({{\textbf{W}}},\textbf{v})       &\nabla_{\bar{\textbf{w}}_2}\nabla_{\bar{\textbf{w}}^C_2} \mathcal{L}({{\textbf{W}}},\textbf{v})
\end{bmatrix} 
$$
and we observe that 
\begin{align*}
&\nabla_{\bar{\textbf{w}}_1}\nabla_{\bar{\textbf{w}}^C_1} \mathcal{L}({{\textbf{W}}},\textbf{v})  = \nabla_{\bar{\textbf{w}}_1}\nabla_{\bar{\textbf{w}}^C_2} \mathcal{L}({{\textbf{W}}},\textbf{v}) =
    \nabla_{\bar{\textbf{w}}_2}\nabla_{\bar{\textbf{w}}^C_1} \mathcal{L}({{\textbf{W}}},\textbf{v})     =   \nabla_{\bar{\textbf{w}}_2}\nabla_{\bar{\textbf{w}}^C_2} \mathcal{L}({{\textbf{W}}},\textbf{v}) \\
&= \frac{1}{2n}\sum^n_{i=1} v_1^*v_1 \psi^{'}(\langle \textbf{w}_1, \textbf{x}_i \rangle)^*\psi^{'}(\langle \textbf{w}_1,\textbf{x}_i \rangle) \textbf{x}_i^C \textbf{x}_i^T \\
& =
 \begin{bmatrix}
    \frac{5}{216} & \frac{1}{216} \\
    \frac{1}{216}      & \frac{5}{216}
\end{bmatrix}.
\end{align*}
The other two are slightly different under Wirtinger calculus. By Wirtinger calculus, we have Wirtinger Hessian
\begin{align*}
\nabla^2_{\bar{\textbf{W}}} \mathcal{L}({{\textbf{W}}}) =
 \begin{bmatrix}
     \frac{1}{2n}\sum^n_{i=1} \mathcal{L}^{*}_i(\textbf{W},\textbf{v})  \frac{\partial^2}{\partial \bar{\textbf{w}}^2_1} \mathcal{L}_i(\textbf{W},\textbf{v}) &  \frac{1}{2n}\sum^n_{i=1} \mathcal{L}^{*}_i(\textbf{W},\textbf{v})  \frac{\partial^2}{\partial \bar{\textbf{w}}_1 \bar{\textbf{w}}_2} \mathcal{L}_i(\textbf{W},\textbf{v}) \\
      \frac{1}{2n}\sum^n_{i=1} \mathcal{L}^{*}_i(\textbf{W},\textbf{v})  \frac{\partial^2}{\partial \bar{\textbf{w}}_2 \bar{\textbf{w}}_1} \mathcal{L}_i(\textbf{W},\textbf{v})     &  \frac{1}{2n}\sum^n_{i=1} \mathcal{L}^{*}_i(\textbf{W},\textbf{v})  \frac{\partial^2}{\partial \bar{\textbf{w}}^2_2} \mathcal{L}_i(\textbf{W},\textbf{v})
\end{bmatrix}.
\end{align*}
We also have $$ \frac{1}{2n}\sum^n_{i=1} \mathcal{L}^{*}_i(\textbf{W},\textbf{v})  \frac{\partial^2}{\partial \bar{\textbf{w}}_1 \bar{\textbf{w}}_2} \mathcal{L}_i(\textbf{W},\textbf{v}) =
      \frac{1}{2n}\sum^n_{i=1} \mathcal{L}^{*}_i(\textbf{W},\textbf{v})  \frac{\partial^2}{\partial \bar{\textbf{w}}_2 \bar{\textbf{w}}_1} \mathcal{L}_i(\textbf{W},\textbf{v}) =0$$ and 
\begin{align*}
&\frac{1}{2n}\sum^n_{i=1} \mathcal{L}^{*}_i(\textbf{W},\textbf{v})  \frac{\partial^2}{\partial \bar{\textbf{w}}^2_1} \mathcal{L}_i(\textbf{W},\textbf{v}) \\
&= \frac{1}{2n}\sum^n_{i=1} (\textbf{v}^T\psi(\textbf{W}\textbf{x}_i) - y_i) v_1 \psi^{''}(\langle \textbf{w}_1, \textbf{x}_i \rangle)\textbf{x}_i\textbf{x}_i^T\\
&=
 \begin{bmatrix}
    \frac{1}{108} & -\frac{1}{108} \\
     -\frac{1}{108} &  \frac{1}{108}
\end{bmatrix}\\
&= \frac{1}{2n}\sum^n_{i=1} \mathcal{L}^{*}_i(\textbf{W},\textbf{v})  \frac{\partial^2}{\partial \bar{\textbf{w}}^2_2} \mathcal{L}_i(\textbf{W},\textbf{v}).
\end{align*}
It can be verified that $ \mathcal{H}^{\mathbb{C}}_{\bar{\textbf{W}}} $ has 1 negative eigenvalues.

\section{Proof of Lemma \ref{lem_24}}
\label{proof_lem_24}

To select $(\hat{\textbf{W}},\hat{\textbf{v}})$ in an arbitrarily small neightbor of $(\bar{\textbf{W}},\bar{\textbf{v}})$, we permute $\bar{\textbf{W}}$ only and let $\hat{\textbf{v}}= \bar{\textbf{v}}$. For an arbitrarily large $N \in \mathbb{N}^{+}$, we let
$$
\hat{\textbf{W}} = 
\begin{bmatrix}
    1-\frac{1}{10^N}  & 1+ \frac{i}{10^N} \\
    1-\frac{1}{10^N}  & 1+ \frac{i}{10^N}
\end{bmatrix}.
$$
Firstly, we notice that $\mathcal{L}(\bar{\textbf{W}},\bar{\textbf{v}}) = \frac{1}{9}$. Therefore, it is enough to show that $\mathcal{L}(\hat{\textbf{W}},\hat{\textbf{v}}) <  \frac{1}{9}$ for an arbitrarily large $N$. By simple calculations, we have
$$
\hat{\textbf{W}} \textbf{X} = 
\begin{bmatrix}
    1-\frac{1}{10^N}   & 1+ \frac{i}{10^N} & 1-\frac{1}{2 \cdot 10^N} + \frac{i}{2 \cdot 10^N} \\
    1-\frac{1}{10^N}       & 1+ \frac{i}{10^N} & 1-\frac{1}{2 \cdot 10^N} + \frac{i}{2 \cdot 10^N}
\end{bmatrix} ,
$$
$$
\psi(\hat{\textbf{W}} \textbf{X}) = 
\begin{bmatrix}
    \frac{(10^N - 1)^2}{10^{2N}}   & \frac{(10^{N} + i)^2}{10^{2N}} & \frac{(2 \cdot 10^N -1 + i)^2}{4 \cdot 10^{2N}} \\
    \frac{(10^N - 1)^2}{10^{2N}}        &  \frac{(10^{N} + i)^2}{10^{2N}} &  \frac{(2 \cdot 10^N -1 + i)^2}{4 \cdot 10^{2N}}
\end{bmatrix} ,
$$
$$
\hat{\textbf{v}} \psi(\hat{\textbf{W}} \textbf{X}) = 
\begin{bmatrix}
    \frac{(10^N - 1)^2}{3 \cdot 10^{2N}}   & \frac{(10^{N} + i)^2}{3 \cdot 10^{2N}} & \frac{(2 \cdot 10^N -1 + i)^2}{12 \cdot 10^{2N}} 
\end{bmatrix}.
$$
Therefore, 
\begin{align*} 
\mathcal{L} ( \hat{\textbf{W}},\hat{\textbf{v}}) &= \frac{1}{6} \cdot \left(\frac{(10^N - 1)^4}{9 \cdot 10^{4N}} + \left\|\frac{(10^N + i)^2}{3 \cdot 10^{2N}}\right\|^2 + \left\| 1-\frac{(2 \cdot 10^N -1 + i)^2}{12 \cdot 10^{2N}} \right\|^2 \right) \\
& = \frac{1}{6} \cdot \left(\frac{(10^N - 1)^4}{9 \cdot 10^{4N}} + \frac{(10^{2N} - 1 )^2}{9 \cdot 10^{4N}} + \frac{4 \cdot10^{2N} }{9 \cdot 10^{4N}} + \frac{(8 \cdot 10^{2N} - 4 \cdot 10^N)^2}{144 \cdot 10^{4N}} +\frac{4 \cdot (2 \cdot 10^{N} - 1)^2}{144 \cdot 10^{4N}} \right)\\
& = \frac{1}{6} \cdot \left( \frac{96\cdot 10^{4N} -128\cdot 10^{3N} +160\cdot 10^{2N} -80\cdot 10^N +36}{144 \cdot 10^{4N}} \right) \\
& < \frac{1}{9}
\end{align*} 
for all $N \in \mathbb{N}^+$.

\section{Additional Preliminaries (Supplement to Section \ref{sec_pre})}

\subsection{More on complex analysis}

We provide some basic definitions of univariate complex functions. The generalization of multivariate functions is the same as in the real case. 

Let $f : \mathbb{C} \mapsto \mathbb{C}$ given by $f(z) = u(z) + iv(z)$ where $z = x+iy$.

\begin{definition} Suppose that $f$ is defined on some open neighbourhood of $z_0$. Then, the derivative of $f$ at $z_0$ is given by
$$
f^{'}(z_0) = \underset{\Delta z \rightarrow 0}{\text{lim}} \frac{f(z_0 + \Delta z) - f(z_0)}{\Delta z}
$$
where $\Delta z = \Delta x + i \Delta y$, provided this limit exists. Such an $f$ is said to be differentiable at $z_0$.
\end{definition}

\begin{definition}[Analytic functions] A complex function $f(z)$ is called analytic at the point $z_0$ if it is differentiable at $z_0$ and in a neighbourhood of $z_0$.
\end{definition}

Some examples of analytic functions include all polynomials, trigonometric functions, and exponential functions.

\begin{definition}[Cauchy-Riemann equations] If $f^{'}(z)$ exists, the partials of $u$ and $v$ exist at $(x,y)$ and satisfy the Cauchy-Riemann equations
$$
\frac{\partial u}{\partial x}(x,y) = \frac{\partial v}{\partial y}(x,y) \text{ and } \frac{\partial u}{\partial y}(x,y) = -\frac{\partial v}{\partial x}(x,y).
$$
\end{definition}

\begin{theorem}[Necessary conditions for differentiability] Suppose that $f$ is differentiable at $z$. Then the Cauchy-Riemann equations hold at $z$ and $f^{'}(z) = \frac{\partial u}{\partial x}(x,y) + i \frac{\partial v}{\partial x}(x,y) =  \frac{\partial v}{\partial y}(x,y) - i  \frac{\partial u}{\partial y}(x,y)$.
\end{theorem}

\begin{theorem}[Sufficient conditions for differentiability] Suppose $f(z)$ is defined throughout some open neighbourhood $U$ of the point $z_0 = x_0 + i y_0$, and suppose that $\frac{\partial u}{\partial x}, \frac{\partial u}{\partial y}, \frac{\partial v}{\partial x}, \frac{\partial v}{\partial y} $ exist everywhere in $U$. Then, if $\frac{\partial u}{\partial x}, \frac{\partial u}{\partial y}, \frac{\partial v}{\partial x}, \frac{\partial v}{\partial y} $ are continuous at $(x_0, y_0)$ and satisfy the Cauchy-Riemann equations at $(x_0, y_0)$, then $f$ is differentiable at $z_0$ and $f^{'}(z) = \frac{\partial u}{\partial x}(x,y) + i \frac{\partial v}{\partial x}(x,y) =  \frac{\partial v}{\partial y}(x,y) - i  \frac{\partial u}{\partial y}(x,y)$.
\end{theorem}

Let $z^*$ denotes the conjugate of $z$ and $|z|$ denotes the modulus of $z$. We recall some properties of complex numbers. For all $z, y \in \mathbb{C}$, we have $|z^*| = |z|, zz^* = |z|^2, z^{-1} = \frac{z^*}{|z|^2} \text{ if } z \neq 0, \mathcal{R}(z) = \frac{z + z^*}{2}, \mathcal{I}(z) = \frac{z - z^*}{2i}, |zy| = |z||y|,$ and $|z^n| = |z|^n.$

\subsection{More on Wirtinger calculus}

We provide few important exposition of Wirtinger calculus here. More explanations can be found in \cite{KreutzDelgado2005TheCG} and \cite{BP10}. Consider the complex-valued function $f : \mathbb{C}^n \mapsto \mathbb{C}$, $f(\textbf{z}) = u(\textbf{x},\textbf{y})+iv(\textbf{x},\textbf{y})$. The Wirtinger derivative and the conjugate Wirtinger derivative are defined to be
$$
\frac{\partial f}{\partial \textbf{z}} := \left[\frac{\partial f}{\partial z_1}, \dots, \frac{\partial f}{\partial z_n}\right], \ \frac{\partial f}{\partial \textbf{z}^C} := \left[\frac{\partial f}{\partial z_1^*}, \dots, \frac{\partial f}{\partial z_n^*}\right]
$$
where
$$
\frac{\partial f}{\partial z_j} := \frac{1}{2}\left(\frac{\partial f}{\partial x_j} - i \frac{\partial f}{\partial y_j}\right) = \frac{1}{2}\left(\frac{\partial u}{\partial x_j} + \frac{\partial v}{\partial y_j}\right) + \frac{i}{2}\left(\frac{\partial v}{\partial x_j} - \frac{\partial u}{\partial y_j}\right),
$$
$$
\frac{\partial f}{\partial z_j^*} := \frac{1}{2}\left(\frac{\partial f}{\partial x_j} + i \frac{\partial f}{\partial y_j}\right) = \frac{1}{2}\left(\frac{\partial u}{\partial x_j} - \frac{\partial v}{\partial y_j}\right) + \frac{i}{2}\left(\frac{\partial v}{\partial x_j} + \frac{\partial u}{\partial y_j}\right).
$$
Note that the Wirtinger derivative is well defined as long as the real functions $u$ and $v$ are differentiable with respect to $\textbf{x}$ and $\textbf{y}$. In our case, the loss function $\mathcal{L}(\textbf{W})$ has well-defined Wirtinger derivative. 

We now have the following lemma which follows directly from the definitions.

\begin{lemma} If f is complex differentiable, then its Wirtinger derivative is the same as the normal derivative, while the conjugate Wirtinger derivative is equal to zero.
$$
\frac{\partial f}{\partial \textbf{z}} = f^{'},\  \frac{\partial f}{\partial \textbf{z}^C} = \textbf{0}.
$$ 
{\it Similarly, if f is conjugate-complex differentiable, then its conjugate Wirtinger derivative is equal to the normal conjugate-complex derivative, while the Wirtinger derivative is equal to zero.}
$$
\frac{\partial f}{\partial \textbf{z}^C} = f^{'}_{*},\  \frac{\partial f}{\partial \textbf{z}} = \textbf{0}.
$$ 
\end{lemma}


We provide expressions for Wirtinger gradient, Wirtinger Hessian, and the second order Taylor's expansion formula,
$$
\widetilde{\nabla} f(\textbf{z}) = \left[ \frac{\partial f}{\partial \textbf{z}}, \frac{\partial f}{\partial \textbf{z}^C} \right]^*,
$$

\begin{align*}
 \widetilde{\nabla}^2 f(\textbf{z}) = 
\begin{pmatrix}
\frac{\partial}{\partial \textbf{z}} (\frac{\partial f}{\partial \textbf{z}})^* &  \frac{\partial}{\partial \textbf{z}^C} (\frac{\partial f}{\partial \textbf{z}})^* \\
\frac{\partial}{\partial \textbf{z}} (\frac{\partial f}{\partial \textbf{z}^C})^* & \frac{\partial}{\partial \textbf{z}^C} (\frac{\partial f}{\partial \textbf{z}^C})^*
\end{pmatrix},
\end{align*}

\begin{align*}
f(\textbf{z} + \textbf{h}) = f(\textbf{z}) + (\widetilde{\nabla} f(\textbf{z}))^*\cdot 
\begin{pmatrix}
\textbf{h} \\
\textbf{h}^C
\end{pmatrix} +
\frac{1}{2} (\textbf{h}^*,\textbf{h}^T)\cdot  \widetilde{\nabla}^2 f(\textbf{z}) \cdot
\begin{pmatrix}
\textbf{h} \\
\textbf{h}^C
\end{pmatrix} + o(||\textbf{h}||^2).
\end{align*}

A point $\textbf{z}$ is called a critical point of $f$ if and only if $\widetilde{\nabla} f(\textbf{z}) = \textbf{0}$. Since the loss function we will be analyzing is real-valued, as in the standard setting, if $\textbf{W}$
is a local minimum of $\mathcal{L}(\textbf{W}) $, then the Wirtinger's Hessian of $\mathcal{L}(\textbf{W})$ is positive semi-definite. 

Lastly we state some important propositions.

\begin{definition}[Conjugate Cauchy Riemann conditions (CCRC)]
$$
\frac{\partial u}{\partial \textbf{x}} = -\frac{\partial v}{\partial \textbf{y}} \text{ and } \frac{\partial u}{\partial \textbf{y}} = \frac{\partial v}{\partial \textbf{x}}
$$
\end{definition}

\begin{proposition} If $f$ is differentiable in the real sense at $(\textbf{x}, \textbf{y})$ and the CCRC hold, then $f$ is conjugate-complex differentiable.
\end{proposition}

\begin{proposition} If $f$ is conjugate-complex differentiable at $\textbf{z}$ then $u$ and $v$ are differentiable in the real sense and they satisfy the conjugate Cauchy Riemann conditions.
\end{proposition}

\begin{proposition} If $f$ is differentiable in the real sense, then
$$
(\frac{\partial f}{\partial \textbf{z}})^C = \frac{\partial f^C}{\partial \textbf{z}^C} \text{ and } (\frac{\partial f}{\partial \textbf{z}^C})^C = \frac{\partial f^C}{\partial \textbf{z}}.
$$
\end{proposition}

Wirtinger derivative share many properties as normal derivatives like linearity, product rule, and chain rule. 

\begin{proposition}[Linearity] If $f, g$ are differentiable in the real sense and $\alpha, \beta \in \mathbb{C}$, then
$$
\frac{\partial (\alpha f + \beta g )}{\partial \textbf{z}} = \alpha \frac{\partial f}{\partial \textbf{z}} + \beta \frac{\partial g}{\partial \textbf{z}},
$$
$$
\frac{\partial (\alpha f + \beta g )}{\partial \textbf{z}^C} = \alpha \frac{\partial f}{\partial \textbf{z}^C} + \beta \frac{\partial g}{\partial \textbf{z}^C}.
$$
\end{proposition}

\begin{proposition}[Product Rule] If $f, g$ are differentiable in the real sense, then
$$
\frac{\partial ( f \cdot  g )}{\partial \textbf{z}} =  \frac{\partial f}{\partial \textbf{z}}g +  \frac{\partial g}{\partial \textbf{z}}f,
$$
$$
\frac{\partial ( f \cdot  g )}{\partial \textbf{z}^C} =  \frac{\partial f}{\partial \textbf{z}^C}g +  \frac{\partial g}{\partial \textbf{z}^C}f.
$$
\end{proposition}

\begin{proposition}[Chain Rule] If $f, g$ are differentiable in the real sense, then
$$
\frac{\partial ( f \circ  g )}{\partial \textbf{z}} =  \frac{\partial f}{\partial \textbf{z}}(g)\frac{\partial g}{\partial \textbf{z}} +  \frac{\partial f}{\partial \textbf{z}^C}(f)\frac{\partial g^C}{\partial \textbf{z}},
$$
$$
\frac{\partial ( f \circ  g )}{\partial \textbf{z}^C} =  \frac{\partial f}{\partial \textbf{z}}(g)\frac{\partial g}{\partial \textbf{z}^C} +  \frac{\partial f}{\partial \textbf{z}^C}(f)\frac{\partial g^C}{\partial \textbf{z}^C}.
$$
\end{proposition}

\end{document}